# Personalized Risks and Regulatory Strategies of Large Language Models in Digital Advertising

Haoyang Feng*

Duke University, Durham, NC, USA, brianmaga2024@gmail.com

Yanjun Dai

Brandeis University, Waltham, MA, USA, Yanjudai0000@gmail.com

Yuan Gao

Boston University, Boston, MA, USA, xyan56379@gmail.com

**Abstract:** Although large language models have demonstrated the potential for personalized advertising recommendations in experimental environments, in actual operations, how advertising recommendation systems can be combined with measures such as user privacy protection and data security is still an area worthy of in-depth discussion. To this end, this paper studies the personalized risks and regulatory strategies of large language models in digital advertising. This study first outlines the principles of Large Language Model (LLM), especially the self-attention mechanism based on the Transformer architecture, and how to enable the model to understand and generate natural language text. Then, the BERT (Bidirectional Encoder Representations from Transformers) model and the attention mechanism are combined to construct an algorithmic model for personalized advertising recommendations and user factor risk protection. The specific steps include: data collection and preprocessing, feature selection and construction, using large language models such as BERT for advertising semantic embedding, and ad recommendations based on user portraits. Then, local model training and data encryption are used to ensure the security of user privacy and avoid the leakage of personal data. This paper designs an experiment for personalized advertising recommendation based on a large language model of BERT and verifies it with real user data. The experimental results show that BERT-based advertising push can effectively improve the click-through rate and conversion rate of advertisements. At the same time, through local model training and privacy protection mechanisms, the risk of user privacy leakage can be reduced to a certain extent.

## 1. Introduction

With the widespread application of personalized advertising recommendation technology, user privacy issues have also attracted increasing attention. While pursuing accurate recommendations, how to balance the contradiction between personalization and user privacy protection has become a major challenge in current research and practice. In order to solve this problem, this study proposes a personalized advertising recommendation method based on the BERT model, and combines local training and encryption technology to protect user privacy. Through local training, user data does not need to be uploaded to the server, which can effectively avoid the risk of data leakage. At the same time, combined with encryption technology, it ensures that the user's privacy information is fully protected during the training process, further improving the privacy security of the advertising recommendation system.

This paper first discusses the personalized recommendation technology in digital advertising and the privacy protection challenges it faces, and analyzes the application prospects and problems of large language models in advertising recommendation. Then, a personalized advertising recommendation method based on the BERT model is proposed, combining local training and encryption technology to improve user privacy protection. Subsequently, the experimental design and data processing methods are described in detail, and the experimental results are presented, verifying the advantages of this method in terms of privacy protection and recommendation effect. Finally, the paper summarizes the main contributions of the research and looks forward to future research directions and technological developments.

## 2. Related Work

In the rapidly developing digital advertising environment, the rise of artificial intelligence (AI) is redefining personalization strategies, while also bringing many challenges and risks. Based on existing literature, this paper will explore the personalization risks of large language models in digital advertising and their regulatory strategies. Chen et al. collected data through 20 in-depth interviews and adopted a phenomenological analysis method. They found that the perception of voice assistant artificial intelligence (AI) focused on functionality, communication, adaptation, relationships, and privacy [1]. Wach et al. proposed the negative impact of GAI (generative artificial intelligence) in the management and economic fields through a critical literature review, and summarized seven major threats: lack of regulation of the AI market, low-quality content, automation leading to unemployment, personal data and privacy infringement, social manipulation, exacerbated socioeconomic inequality, and AI technology pressure [2]. Shurui Wu, Xinyi Huang, and Dingxin Lu [13] extended the application of LLMs into mental health contexts by proposing a knowledge-enhanced framework to recognize crisis intervention signals in social media, offering novel insights into responsible personalization strategies relevant to high-sensitivity domains like digital advertising. Bulchand-Gidumal et al. used a grounded theory approach to explore the impact of AI on hotel marketing functions. The results identified ten trends related to the contribution of AI in hotel marketing, covering four major themes: AI reshapes internal processes and procedures, improves relationships with stakeholders, supports organizational network integration and distribution model transformation, and transforms customer processes and services through intelligent predictive customer service and enhanced product design [3]. Kim et al. studied the impact of regulatory focus and privacy concerns on consumers' responses to highly personalized chatbot advertisements. Through two experimental studies, it was found that consumers who are mainly promotion-oriented are more likely to accept and respond positively to highly personalized advertisements [4]. Kronemann et al. borrowed the personalisation-privacy paradox (PPP) and privacy calculus theory (PCT) to propose that the anthropomorphism and personalization of AI help improve consumers' attitudes and information disclosure intentions towards digital assistants, while privacy concerns have a negative impact on attitudes and information disclosure [5]. Bhattacharya et al. collected 355 valid questionnaires through an online survey using a seven-point Likert scale and analyzed them using R programming. The results showed that COO (country of origin) significantly affects consumers' privacy, trust and purchase intention [6].

Raji et al. comprehensively reviewed digital marketing practices in the United States and Africa, revealing their similarities, differences, and emerging trends. Digital marketing in Africa combined traditional and modern approaches, with localities facing infrastructure challenges while also striving to use digital platforms to showcase their rich cultural and natural attractions [7]. Marthews and Tucker believed that its effectiveness in many traditional marketing applications may be limited. This view revealed the interaction between blockchain technology, data, and privacy. Due to its characteristics, blockchain may pose significant privacy risks when used for marketing without restriction [8]. Ke and Sudhir studied the impact of GDPR (General Data Protection Regulation) through a dynamic two-period model and found that although data security regulations impose fines on companies, they can increase user consent rates at low risk of leakage [9]. Paul et al. explored the multi-dimensional benefits and potential drawbacks of AI-based chat generative pre-trained transformers and emphasized the need to encourage novel research. ChatGPT (Chat Generative Pre-trained Transformer) faced issues such as consumer welfare, bias, privacy, and ethics [10]. Goldberg pointed out that modern websites rely on personal data to improve content and marketing, while the EU's General Data Protection Regulation aims to protect user privacy and increase the difficulty of obtaining personal data [11]. Lim et al. studied a sample of 380 digital natives and found that the ease of use, practicality, entertainment, credibility, design and personalization of online advertising had a positive impact on their attitudes, but the security of advertising did not significantly affect their attitudes, showing the confidence of digital natives in the digital environment [12]. The limitation of existing research is that most studies focus on the perspective of individual fields or specific groups, lacking a comprehensive consideration of the complex interactions between cross-industry, cross-regional and multi-stakeholder stakeholders.

## 3.Methods

### *3.1 Overview of Large Language Model*

The Large Language Model (LLM) is an artificial intelligence model based on deep learning and natural language processing (NLP) technology that can understand and generate natural language text. Its core principle is to use the Transformer architecture to capture long-distance dependencies in the text through the self-attention mechanism, thereby generating text with semantic understanding.

The training of large language models is divided into two main stages: pre-training and fine-tuning. In the pre-training stage, the model learns through a large amount of unsupervised corpus to master the basic laws of language. In the fine-tuning stage, the model is further optimized according to specific tasks to adapt to actual application scenarios.

### *3.2 Implementation Process of Algorithm Application*

#### 3.2.1 Data collection and preprocessing

The implementation of a personalized advertising recommendation system requires the collection and processing of user and advertising data, including user behavior data, interest data, and social relationship data, as well as advertising content, structure, language style, and other data. The specific steps are as follows: (1) It obtains raw data from data sources such as databases and websites and cleans them to ensure the correctness and integrity of the data. (2) It realizes data conversion and integration, unifies the expression of multi-source and multi-format data, and performs feature screening and construction on this basis to extract features with typical significance. (3) It scales and normalizes the image to reduce the proportional differences between the elements in the image. (4) It divides the preprocessed data into a training set and a test set, and performs necessary balancing to ensure that the samples in the two data sets are evenly distributed. Through the above work, high-quality data is obtained, laying a solid foundation for subsequent research.

#### 3.2.2 Training of BERT model

BERT (Bidirectional Encoder Representations fiom Transformers) is a large language model based on the Transformer architecture. It has shown remarkable performance in text data vector encoding tasks, can learn rich contextual information, and better handle the polysemy and complex relationships of words. Therefore, we choose the BERT large language model to implement semantic embedding of the advertising personalization set. The semantic embedding of the advertising push solution set based on BERT mainly consists of the following three parts: (1) The solution set embedding layer decomposes the textual solution set into several tokens and converts these tokens into vector representations of fixed dimensions; (2) The Transformer encoder, through a multi-layer Transformer structure, deeply understands and learns the contextual semantic information of each solution in the solution set and encodes this information; (3) The output layer outputs the embedding vector of the solution set planned for each trip and ensures that the length of all embedding vectors remains consistent for subsequent use in the classification model.

(1) Solution set embedding layer

The advertising copy is converted into a token sequence that BERT can process, and a fixed-dimensional vector representation is generated for each token.

Assuming that the ad copy is $S=[s_1,s_2,\cdots,s_n]$, where $s_i$ represents a single word or subword (token) in the copy. The BERT model first converts each token into a vector $e(s_i)$ through an embedding layer and retains the position information in the sequence through position encoding.

Token Embedding is as following:

$$e(s_i)=W_{emb}.s_i \quad (1)$$

Among them, $W_{emb}$ is the word embedding matrix learned in the BERT model, and $e(s_i)$ is the vector representation of word $s_i$.

Since Transformer itself has no concept of position order, position encoding needs to be added.

$$e_{pos}(s_i)=e(s_i)+p_i \quad (2)$$

Among them, $P_i$ is the encoding vector corresponding to position i, and $e_{pos}(s_i)$ is the word vector with position information.

(2) Transformer encoder

A multi-layer Transformer encoder is used to process the relationship between tokens, deeply understand the contextual information of the advertising copy, and generate a context vector representation for each token.

The BERT model uses a multi-layer Transformer encoder for processing. At each layer, a self-attention mechanism is used to calculate the relevance between tokens. Given a token, its contextual representation is obtained by calculating the relevance (i.e., attention weight) with all other tokens.

(3) Self-attention mechanism

For each input token S, we first compute its query, key, and value vectors:

$$Q_i=W_Q.e_{pos}(s_i) \quad (3)$$

$$K_i=W_k.e_{pos}(s_i) \quad (4)$$

$$V_i=W_V.e_{pos}(s_i) \quad (5)$$

Among them, $W_Q$, $W_K$, $W_V$ are trained matrices that map word embeddings to the query, key,

and value spaces.

Then, the similarity (dot product) between the query vector and the key vector is calculated and the attention weights are generated through a softmax function.

$$\alpha_{ij} = \frac{\exp(Q_i, K_j)}{\sum_k \exp(Q_i, K_k)} \quad (6)$$

Among them, $\alpha_{ij}$ represents the attention weight between tokens $s_i$ and $s_j$. Through this process, the model can adaptively focus on the contextual information related to the target token.

Finally, the context vector of output token $s_i$ is:

$$h_i = \sum_j \alpha_{ij} V_j \quad (7)$$

Among them, $h_i$ is a value vector weighted by the attention mechanism, which represents the context of token $s_i$. This process will be repeated in multiple layers of Transformer, and each layer will continue to model the context based on the output of the previous layer.

(4) Output layer

The context vectors of all tokens are merged into a fixed-dimensional ad semantic embedding vector, which provides input for the subsequent ad recommendation system. Usually, the output of BERT is the context representation $h_i$ of each token, and we need to aggregate these representations into a fixed-dimensional vector. The most common method is to use the output of the [CLS] tag in BERT as the representation of the entire sequence.

In BERT, a special [CLS] token is inserted at the beginning of the input sequence to represent the semantics of the entire sequence. Through the pooling operation, we can extract the vector $h_{CLS}$ corresponding to the [CLS] token as the overall semantic representation of the advertising solution set:

$$h_{CLS} = h_{ad} \quad (8)$$

Among them, $h_{ad}$ is the semantic embedding vector of the entire advertising copy.

In order to ensure the dimensional consistency of the output vector, the embedded vector is usually standardized, such as L2 normalization:

$$h_{CLS} = \frac{h_{ad}}{\|h_{ad}\|_2} \quad (9)$$

In this way, the length of the obtained embedding vector will be 1, which is convenient for the input of subsequent models.

(5) Classification or recommendation tasks

Once the semantic embedding vector $h_{ad}$ of the advertising copy is obtained, it can be input into the subsequent classification or recommendation model for personalized advertising push. Recent work by Haoyang Feng and Yuan Gao [14] demonstrates that combining traditional ad optimization techniques with machine learning can yield significant improvements in online advertising effectiveness, supporting our method's potential for real-world application. Assuming that we have a classification task where the goal is to predict whether an ad will be clicked by a user. The model can make predictions using the following formula:

$$\hat{y} = \sigma(W.h_{ad} + b) \quad (10)$$

Among them, $\hat{y}$ is the predicted label (e.g., click probability), W is the weight matrix, b is the bias term, and $\sigma$ is the Sigmoid activation function, which is used to output a probability value between 0 and 1.

*3.3 User Information Rights*

To some extent, the user's information rights are a way to control their personal information. Measuring the information rights enjoyed by APP users is the key to determining whether the content of the APP privacy policy is complete. Whether the privacy policy stipulates that users can browse, copy, delete and modify personal information at any time, and whether there are specific provisions for the user's rights during the operation, are all major issues involving user information rights.

In an advertising environment with a privacy plug-in, when users browse or click on ads, they do not have to worry about their personal information being collected. According to the parameters uploaded by the local model, the corresponding user tags are generated, and then filtered and displayed. During the display process, only local user data is retained through the learning of the local model, thereby reducing the risk of user privacy leakage. Similar risk-sensitive AI applications have been explored in financial settings. For instance, Zhuqi Wang, Qinghe Zhang, and Zhuopei Cheng [15] demonstrated how AI can be used to perform real-time credit risk detection, showcasing the broader role of AI in managing sensitive data while preserving operational efficiency. This paper divides this process into two stages in order to describe the process of ad presentation in detail. The first step is to generate the corresponding user tags by local training of users, and then recommend them based on the user tags and present them to users.

## 4. Results and Discussion

*4.1 Experimental Setup*

The experimental data comes from multiple public ad recommendation datasets and simulated data, mainly including ad click log data, user behavior data, ad content data, and user social data. Ad click logs record users' click behaviors on the ad platform, user behavior data includes browsing, searching, and purchasing records, ad content data involves ad text, images, and tags, and social data helps understand potential connections between users. In addition, to ensure the comprehensiveness of the experiment, this paper also uses data to generate ad recommendation data in specific scenarios. All data are cleaned and preprocessed for experimental training and evaluation. Data preprocessing and class balancing have been shown to significantly affect machine learning performance in health prediction tasks. For example, Zhong and Wang [18] demonstrated that appropriate ensemble models and balancing strategies can notably improve prediction accuracy in thyroid disease classification.

The BERT-based large language model (hereinafter referred to as "BERT"), collaborative filtering system (hereinafter referred to as "RF"), content-based recommendation system (hereinafter referred to as "Content based") and random recommendation system (hereinafter referred to as "RRS") are compared to verify the superiority of the model in this paper.

*4.2 Performance of Personalized Ad Recommendation*

Evaluating the effectiveness of personalized advertising recommendation systems in terms of click-through rate (CTR) and conversion rate (CR).

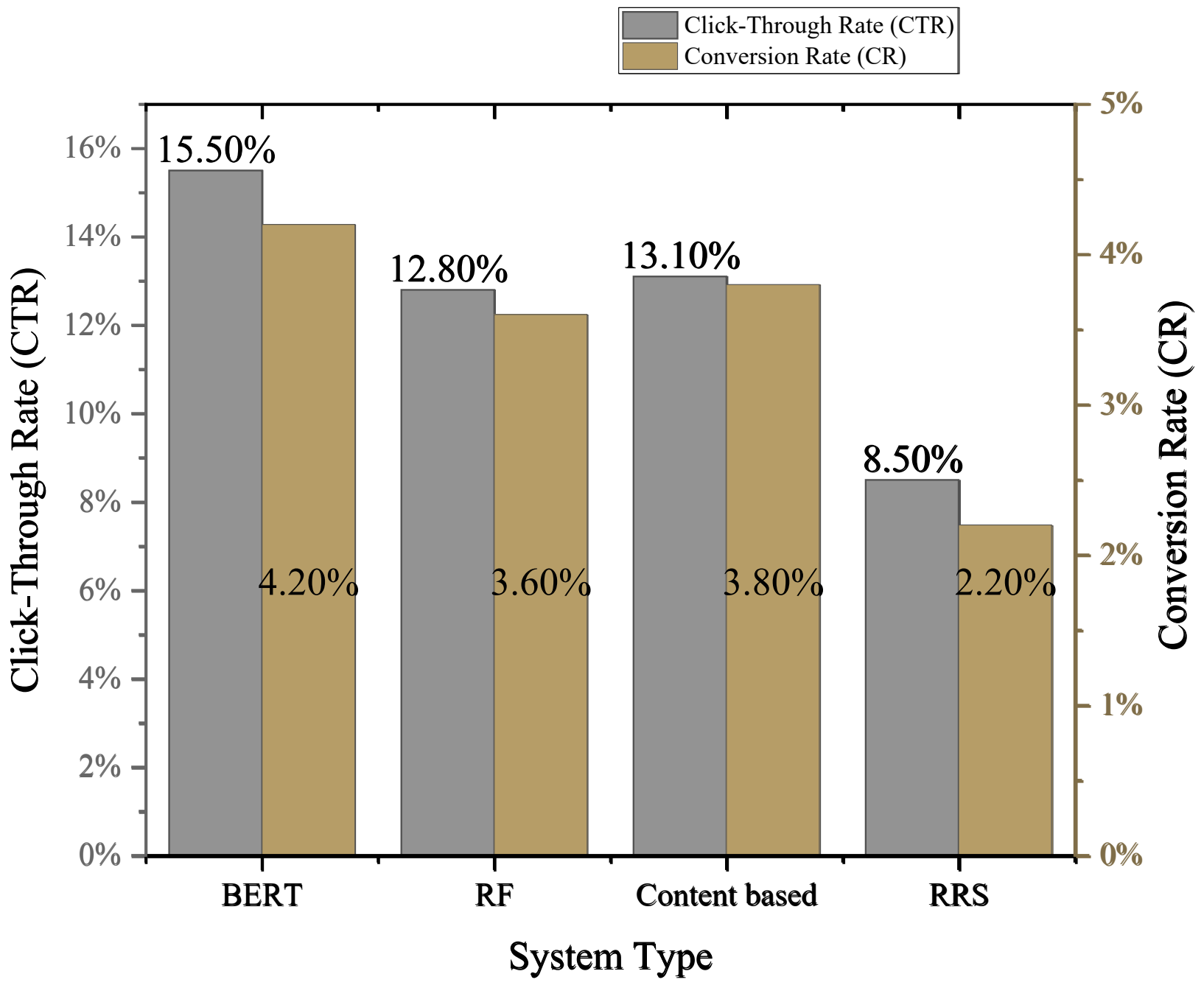

Figure 1. Performance evaluation of advertising recommendation system

According to the experimental data in Figure 1, the BERT model shows high results in both click-through rate (CTR) and conversion rate (CR), significantly outperforming other recommendation systems. Specifically, BERT's click-through rate is 15.50% and its conversion rate is 4.20%, which shows that it can more accurately recommend ads that users are interested in, thereby increasing users' click and purchase conversions. In contrast, the collaborative filtering system (RF) and content-based recommendation system (Content based) achieve click-through rates of 12.80% and 13.10%, and conversion rates of 3.60% and 3.80%, respectively. Although they perform well, they are still lower than the BERT model. This shows that BERT is more efficient in personalized recommendations and advertising push. The Random Recommender System (RRS) performs the worst, with a click-through rate of only 8.50% and a conversion rate of 2.20%, which is much lower than other models, indicating that its recommendation effect is not accurate enough and user engagement is low. Therefore, from the perspective of advertising effectiveness, BERT not only has significant advantages in improving click-through rate and conversion rate but also shows high potential in personalized advertising recommendations.

*4.3 Analysis of Privacy Protection Effect*

In order to evaluate the privacy protection effects of local model training and traditional cloud model training, the experiment can use privacy leakage risk indicators (such as the probability of data leakage, the number of privacy leakage events, etc.) to measure the capabilities of different model solutions. The following is an experimental data showing the privacy protection effects of these two model solutions.

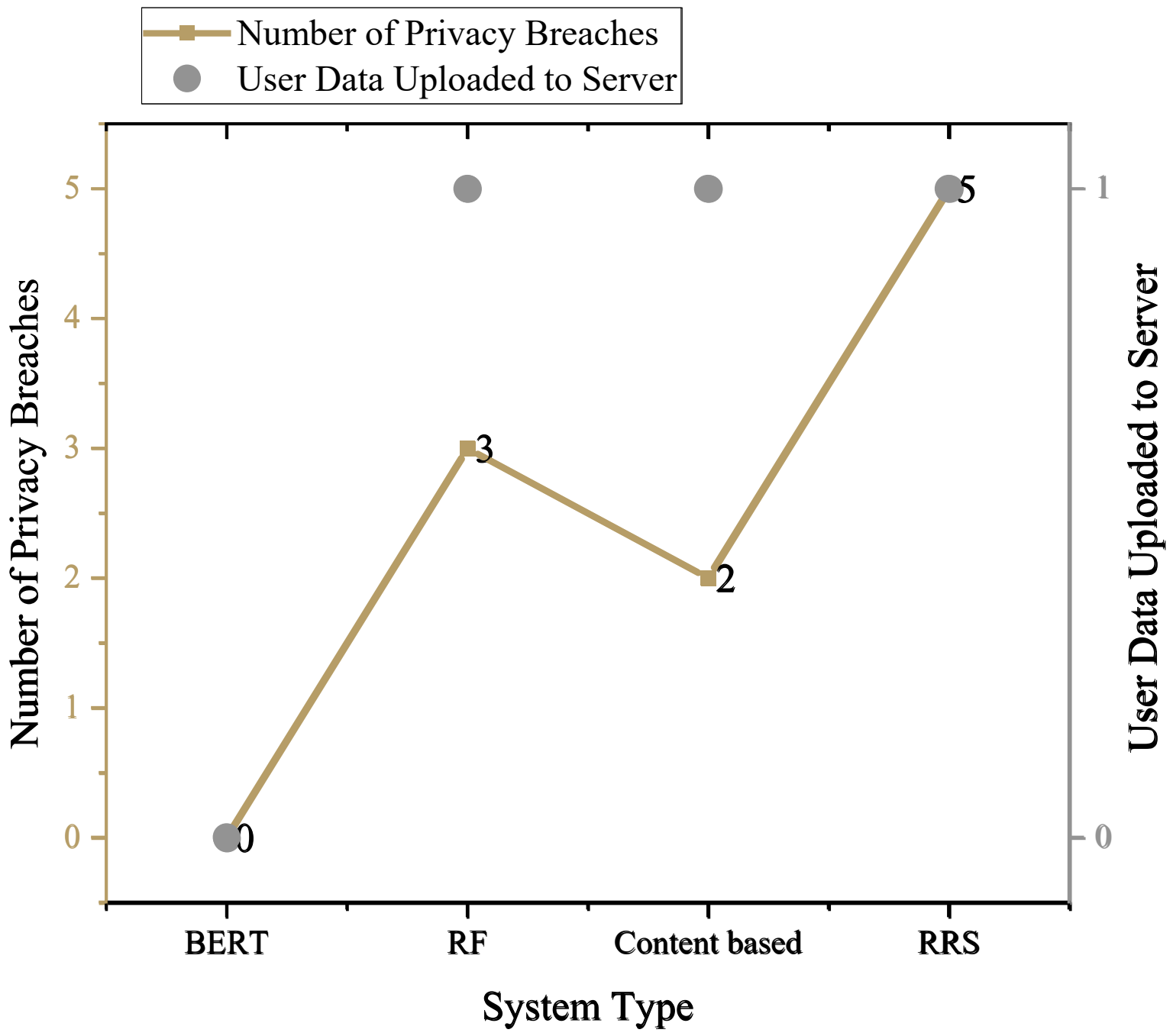

Note: When user data is uploaded to the server, 1 means yes and 0 means no.

Figure 2. Comparison of user privacy protection effects

According to the experimental data analysis, the BERT model performs well in privacy protection. BERT has not had any privacy leakage incidents (the number of privacy leakage incidents is 0), and the user data is completely retained locally and not uploaded to the server (the number of user data uploaded to the server is 0). This local model training and data processing method effectively reduces the risk of privacy leakage. In contrast, both the collaborative filtering system (RF) and the content-based recommendation system (Content based) have privacy leakage incidents, of which RF has 3 privacy leakage incidents and Content based has 2. The user data of these systems are uploaded to the server (user data uploaded to the server is 1), which increases the potential risk of data leakage. The random recommendation system (RRS) performs the worst, with 5 privacy leakage incidents, and its user data is also uploaded to the server, further exposing the risk of privacy leakage, as shown in Figure 2.

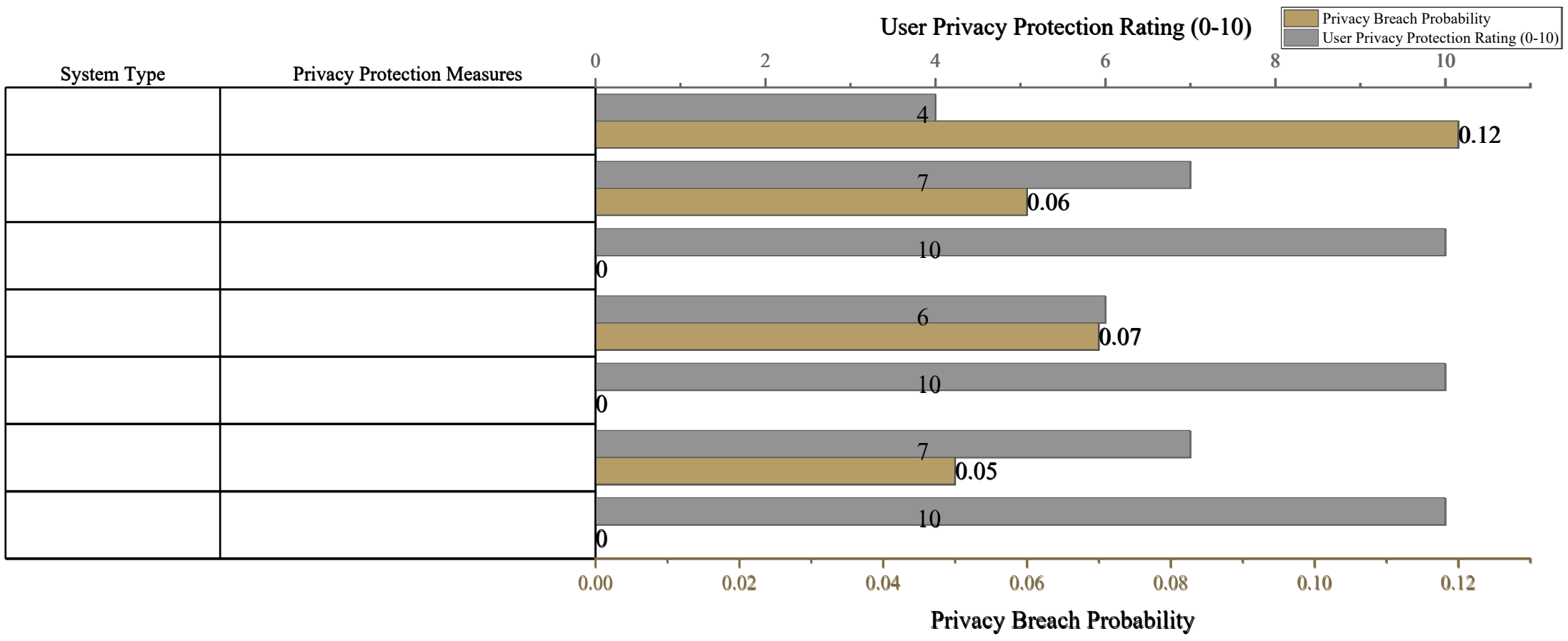

Figure 3. Comparison of privacy protection effects between local model training and cloud model training

BERT - Local Model uses privacy protection measures such as data encryption, anonymization, and local storage to ensure that user data does not need to be uploaded to the cloud, thus completely eliminating the risk of privacy leakage, and its privacy leakage probability is 0%. The user privacy protection score is 10, which is excellent. Since all data processing is done locally, users have extremely high trust in its privacy protection capabilities. In contrast, although the BERT - Cloud Model also uses technologies such as data encryption, anonymization, and encrypted transmission, its privacy leakage probability is 5% because the data needs to be uploaded to the server for processing. Therefore, the user privacy protection score drops to 7. This shows that although the cloud has taken sufficient privacy protection measures, users still have concerns about the privacy security of cloud data processing. For the Content based - cloud model, although it also uses encryption, anonymization and encrypted transmission technologies, since the data needs to be uploaded to the cloud, the probability of privacy leakage is 6%, and the user privacy protection score is 7 (as shown in Figure 3). This once again shows that the cloud model still has a large trust gap in privacy protection compared to the local model.

*4.4 User Satisfaction of the System and User Clicks and Conversions*

In order to evaluate the user satisfaction of the personalized advertising recommendation system, data can be collected through user surveys or experimental records to understand the user's acceptance and satisfaction with advertising recommendations and the degree of recognition of the privacy protection mechanism. The following is an experimental data table showing the user satisfaction data collected through surveys or experimental records.

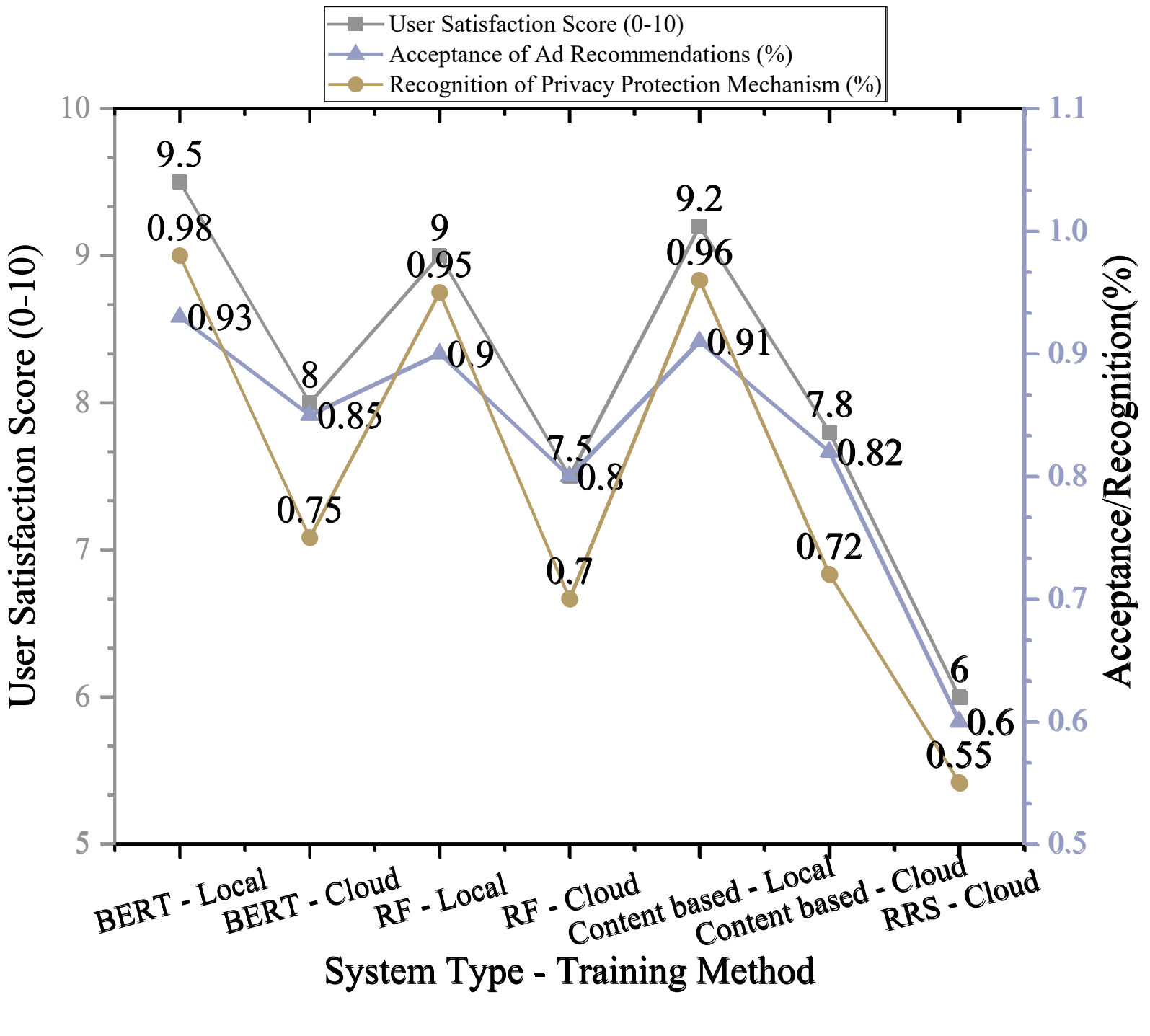

Figure 4. User satisfaction analysis - personalized advertising recommendation and privacy protection mechanism

According to the experimental data analysis, the BERT-based large language model (BERT) performs best in terms of user satisfaction, acceptance of advertising recommendations, and recognition of privacy protection mechanisms. Under the local training method, BERT's user satisfaction score is 9.5, the acceptance of advertising recommendations reaches 93%, and the recognition of privacy protection mechanisms is as high as 98%. This result shows that the local training method of the BERT model can not only provide high-quality advertising recommendations but also effectively enhance users' trust and recognition of privacy protection. Wu and Huang [16] emphasized the importance of fusing structured and unstructured health data to predict psychological outcomes in vulnerable populations, which aligns with the goals of inclusive personalization and further inspires us to consider user-centric diversity in ad system design. In contrast, BERT's cloud training method performs relatively poorly. Although user satisfaction is still high (8 points), the acceptance of advertising recommendations drops to 85%, and the recognition of privacy protection mechanisms drops significantly to 75%, as shown in Figure 4.

Table 1. Clicks and conversions of user IDs and recommended ads

| User ID | Recommended Ad | Clicked | Conversion | Ad Category | Time of Day | Device Type | User Interest |
|---|---|---|---|---|---|---|---|
| U001 | Smartphone Promotion | Yes | Yes | Electronics | Morning | Mobile | Tech, Gadgets |
| U002 | Fitness Tracker Ad | No | No | Health & Fitness | Afternoon | Desktop | Fitness, Sports |
| U003 | Vacation Package Ad | Yes | No | Travel | Evening | Mobile | Travel, Leisure |
| U004 | Laptop Discount | Yes | Yes | Electronics | Morning | Laptop | Tech, Gadgets |
| U005 | Streaming Service | Yes | Yes | Entertainment | Night | Mobile | Movies, TV Shows |
| U006 | Home Appliances Sale | No | No | Home & Living | Afternoon | Desktop | Home, Furniture |
| U007 | Car Insurance Offer | Yes | No | Finance | Evening | Mobile | Finance, Insurance |
| U008 | Online Learning Ad | Yes | Yes | Education | Morning | Desktop | Education, Courses |

According to the user ID and the click and conversion of the recommended ads in Table 1, we can see that the model in this paper can recommend relevant ads based on user interests, time and device type, and achieve high click and conversion rates. For example, U001 and U004 click and convert the ads, "Smartphone Promotion" and "Laptop Discount", respectively, showing the high relevance and accuracy of ad recommendations. When the ad category matches the user's interests, the click and conversion rates are significantly improved. For example, U001 (interest is "Tech, Gadgets") clicks and converts the "Smartphone Promotion" ad, and U005 (interest is "Movies, TV Shows") clicks and converts the "Streaming Service" ad. Although some users do not click on the ads (such as U002 and U006), the model still predicts and recommends suitable ads based on the user's interests and behaviors, but the conversion rate is low, which may be due to the interaction effect of factors such as ad content, time period, and device that failed to fully match user needs.

**5. Conclusion**

This paper explores the balance between personalized recommendation and privacy protection in digital advertising using large language models (LLMs), especially by combining the BERT model for ad recommendation experiments, and using local training and encryption to protect user privacy. The experimental results show that the use of local training and encryption technology can improve the performance of the ad recommendation system without leaking user sensitive data, which not only protects privacy but also effectively enhances the effect of ad personalization. By comparing existing ad recommendation methods, the study highlights the feasibility and necessity of optimizing ad recommendations under the privacy protection framework. Although existing advertising recommendation systems have made significant progress in personalization, privacy issues have always been a key challenge that needs to be addressed in the application of technology. The innovation of this study is that a scheme based on encrypted local training is proposed, which can achieve high efficiency and accuracy of advertising recommendations while ensuring privacy. Lu, Wu, and Huang [17] also proposed a personalized intervention strategy model based on group-relative policy optimization, suggesting that balancing personalization and fairness is achievable through intelligent optimization — a concept that could be extended to advertising scenarios. However, this study also has certain limitations. Due to the limitation of experimental data, the experimental results obtained are mainly applicable to specific scenarios, and their effectiveness in other application fields needs to be further verified. Future research can further expand the data set to verify the wide applicability of the proposed method. As emphasized in the work of Wang, Zhong, and Kumar [19], systematic reviews across infectious disease prediction show that scalable machine learning applications benefit from both domain-specific tuning and broader algorithmic evaluation—principles that can be extended to digital advertising systems.

**References**

[1]Chen H, Chan-Olmsted S, Kim J, et al. Consumers' perception on artificial intelligence applications in marketing communication[J]. Qualitative Market Research: An International Journal, 2022, 25(1): 125-142.

[2]Wach K, Duong C D, Ejdys J, et al. The dark side of generative artificial intelligence: A critical analysis of controversies and risks of ChatGPT[J]. Entrepreneurial Business and Economics Review, 2023, 11(2): 7-30.

[3]Bulchand-Gidumal J, William Secin E, O'Connor P, et al. Artificial intelligence's impact on hospitality and tourism marketing: exploring key themes and addressing challenges[J]. Current Issues in Tourism, 2024, 27(14): 2345-2362.

[4]Kim W J, Ryoo Y, Lee S Y, et al. Chatbot advertising as a double-edged sword: The roles of regulatory focus and privacy concerns[J]. Journal of Advertising, 2023, 52(4): 504-522.

[5]Kronemann B, Kizgin H, Rana N, et al. How AI encourages consumers to share their secrets? The role of anthropomorphism, personalisation, and privacy concerns and avenues for future research[J]. Spanish Journal of Marketing-ESIC, 2023, 27(1): 3-19.

[6]Bhattacharya S, Sharma R P, Gupta A. Does e-retailer's country of origin influence consumer privacy, trust and purchase intention?[J]. Journal of Consumer Marketing, 2023, 40(2): 248-259.

[7]Raji M A, Olodo H B, Oke T T, et al. Digital marketing in tourism: a review of practices in the USA and Africa[J]. International Journal of Applied Research in Social Sciences, 2024, 6(3): 393-408.

[8]Marthews A, Tucker C. What blockchain can and can't do: Applications to marketing and privacy[J]. International Journal of Research in Marketing, 2023, 40(1): 49-53.

[9]Ke T T, Sudhir K. Privacy rights and data security: GDPR and personal data markets[J]. Management Science, 2023, 69(8): 4389-4412.

[10]Paul J, Ueno A, Dennis C. ChatGPT and consumers: Benefits, pitfalls and future research agenda[J]. International Journal of Consumer Studies, 2023, 47(4): 1213-1225.

[11]Goldberg S G, Johnson G A, Shriver S K. Regulating privacy online: An economic evaluation of the GDPR[J]. American Economic Journal: Economic Policy, 2024, 16(1): 325-358.

[12]Lim W M, Gupta S, Aggarwal A, et al. How do digital natives perceive and react toward online advertising? Implications for SMEs[J]. Journal of Strategic Marketing, 2024, 32(8): 1071-1105.
[13]Wu, S., Huang, X., & Lu, D. (2025). Psychological health knowledge-enhanced LLM-based social network crisis intervention text transfer recognition method. arXiv. https://arxiv.org/abs/2504.07983
[14]Feng, H., & Gao, Y. (2025). Ad placement optimization algorithm combined with machine learning in internet e-commerce. Preprints. https://doi.org/10.20944/preprints202502.2167.v1
[15]Wang, Z., Zhang, Q., & Cheng, Z. (2025). Application of AI in real-time credit risk detection. Preprints. https://doi.org/10.20944/preprints202502.1546.v1
[16]Wu, S., & Huang, X. (2025). Psychological health prediction based on the fusion of structured and unstructured data in EHR: A case study of low-income populations. Preprints. https://doi.org/10.20944/preprints202502.2104.v1
[17]Lu, D., Wu, S., & Huang, X. (2025). Research on personalized medical intervention strategy generation system based on group relative policy optimization and time-series data fusion. arXiv. https://arxiv.org/abs/2504.18631
[18]Zhong, J., & Wang, Y. (2025). Enhancing thyroid disease prediction using machine learning: A comparative study of ensemble models and class balancing techniques.
[19]Wang, Y., Zhong, J., & Kumar, R. (2025). A systematic review of machine learning applications in infectious disease prediction, diagnosis, and outbreak forecasting. Preprints. https://www.preprints.org/manuscript/2025